\documentclass[sigconf]{acmart}
\usepackage{amsmath,amssymb,amsfonts}
\usepackage{multirow}
\usepackage{booktabs}
\usepackage{graphicx}
\usepackage{textcomp}
\usepackage{xspace}
\usepackage{CJKutf8}
\usepackage{algorithm}
\usepackage{algpseudocode}
\usepackage{svg}
\usepackage{CJKutf8}
\usepackage{caption}

\usepackage{subfigure}
\usepackage{verbatim}
\usepackage{diagbox}
\usepackage{balance}

\AtBeginDocument{%
  }

\copyrightyear{2025}
\acmYear{2025}
\setcopyright{acmlicensed}\acmConference[CIKM '25]{Proceedings of the 34th
ACM International Conference on Information and Knowledge
Management}{November 10--14, 2025}{Seoul, Republic of Korea}
\acmBooktitle{Proceedings of the 34th ACM International Conference on
Information and Knowledge Management (CIKM '25), November 10--14, 2025,
Seoul, Republic of Korea}
\acmDOI{10.1145/3746252.3761022}
\acmISBN{979-8-4007-2040-6/2025/11}

\newcommand{\ie}{\emph{i.e.,}\xspace}

\newcommand{\baby}{S\textsc{qator}\xspace}

\begin{document}

\captionsetup[figure]{labelfont={bf},name={Fig.},labelsep=period}
\bibliographystyle{ACM-Reference-Format}




\title{Weakly Supervised Fine-grained Span-Level Framework for Chinese Radiology Report Quality Assurance}  



%

\author{Kaiyu Wang}
\authornote{Key Laboratory of Symbolic Computation and Knowledge Engineering of Ministry of Education, Jilin University}
\authornote{Both authors contributed equally to this research.}
\email{kaiyu21@mails.jlu.edu.cn}
\affiliation{
\institution{College of Computer Science and Technology, Jilin University}
\city{Changchun}
\country{China}
}
\author{Lin Mu}
\authornotemark[2]
\email{doctormulin@jlu.edu.cn}
\affiliation{
\institution{Department of Radiology, The First Hospital of Jilin University}
\city{Changchun}
\country{China}
}

\author{Zhiyao Yang}
\authornotemark[1]
\email{zhiyaoy20@mails.jlu.edu.cn}
\affiliation{
\institution{College of Computer Science and Technology, Jilin University}
\city{Changchun}
\country{China}
}

\author{Ximing Li}
\authornotemark[1]
\authornote{Corresponding author}
\email{liximing86@gmail.com}
\affiliation{
\institution{College of Computer Science and Technology, Jilin University}
\city{Changchun}
\country{China}
}

\author{Xiaotang Zhou}
\email{zhouxiaotang@ccut.edu.cn}
\affiliation{
\institution{School of Computer Science and Engineering, Changchun University of Technology}
\city{Changchun}
\country{China}
}

\author{Wanfu Gao}
\authornotemark[1]
\email{gaowf@jlu.edu.cn}
\affiliation{
\institution{College of Computer Science and Technology, Jilin University}
\city{Changchun}
\country{China}
}

\author{Huimao Zhang}
\email{huimao@jlu.edu.cn}
\affiliation{
\institution{Department of Radiology, The First Hospital of Jilin University}
\city{Changchun}
\country{China}
}

\renewcommand{\shortauthors}{Kaiyu Wang et al.}

\begin{abstract}
Quality Assurance (QA) for radiology reports refers to judging whether the junior reports (written by junior doctors) are qualified. The QA scores of one junior report are given by the senior doctor(s) after reviewing the image and junior report. This process requires intensive labor costs for senior doctors.
Additionally, the QA scores may be inaccurate for reasons like diagnosis bias, the ability of senior doctors, and so on.
To address this issue, we propose a Span-level Quality Assurance EvaluaTOR (\baby) to mark QA scores automatically. 
Unlike the common document-level semantic comparison method, we try to analyze the semantic difference by exploring more fine-grained text spans. 
Specifically, \baby measures QA scores by measuring the importance of revised spans between junior and senior reports, and outputs the final QA scores by merging all revised span scores. We evaluate \baby using a collection of 12,013 radiology reports. Experimental results show that \baby can achieve competitive QA scores. 
Moreover, the importance scores of revised spans can be also consistent with the judgments of senior doctors.


\end{abstract}



\maketitle

\section{Introduction}
The rise of the electronic health record (EHR) is generating new challenges and opportunities in the medical domain.
The radiology report, which is one of the formal products of a diagnostic imaging referral in the EHR, is used for communication and documentation purposes.
Radiology report errors occur for many reasons including the use of pre-filled report templates, wrong-word substitution, nonsensical phrases, and missing words. Reports may also contain clinical errors that are not specific to speech recognition including wrong laterality and gender-specific discrepancies. Accordingly, \textbf{Q}uality  \textbf{A}ssurance (QA) for radiology reports becomes a key step of clinical medicine diagnosis, which improves the correction of report errors and thus the accuracy of the clinical record, thereby positively impacting patient management.


In this study, we concentrate on a popular QA pattern commonly applied in clinical medicine diagnosis, especially for hospitals in China \cite{Fu2016Urgent}. Specifically, for each radiographic image, its corresponding report will be finished by at least two doctors, including a junior doctor and senior doctor(s). First, a junior doctor will write an original report, named junior report, by viewing the radiographic image; and then the senior doctor(s) will review this radiographic image, so as to revise the junior report as a new version, named senior report, for assuring the accuracy of the report for the final diagnosis. During this procedure, the senior doctor(s) simultaneously evaluate the quality of the junior report by marking a QA score. QA can be achieved by analyzing a large amount of historical data and providing guidance to residents and teachers, comprehensively improving the quality of the image reports, and implementing standardization in the training process.

In the QA task for radiology reports, the QA score is a key factor, which can be used not only to analyze the characteristics of reports but also as clinical tests for junior doctors. Therefore, we expect the QA scores made by senior doctors to be accurate, objective, standard, and consistent. Unfortunately, due to various causes such as difficulties in diagnosis, ability differences between senior doctors, and diagnosis bias, the QA scores in real diagnosis scenarios eventually tend to be inaccurate, subjective, nonstandard, and inconsistent to some extent. To make matters worse, analyzing radiographic images requires intensive labor for doctors, however, few doctors sometimes have to handle a large number of reports in a limited time, further eventually lowering the quality of QA scores.

To address this issue, we raise the question of whether we can automatically mark the QA scores by using a predictive model, instead of human beings, so as to simultaneously improve the score quality and save many human efforts. To answer this question, we are especially inspired by the techniques of natural language processing (NLP), which have been successfully applied to a variety of healthcare and medical applications, such as clinical decision support \cite{yang2022social,zhangLW23a}, EHR management \cite{jade23using,elias23natural}, and drug discovery \cite{richard18hybird}, to name a few. Technically, a straightforward way is to formulate the QA task as a typical NLP task of semantic textual similarity (STS) \cite{zhang2019bertscore}, whose goal is to predict the difference degrees between any two texts. We can collect a training dataset containing a number of junior and senior report pairs with manual QA scores (as difference degrees), and then use it to induce an STS model that can predict the QA score given any future junior and senior report pair.

\begin{figure}[t]
    \centering
    \includegraphics[scale=0.45]{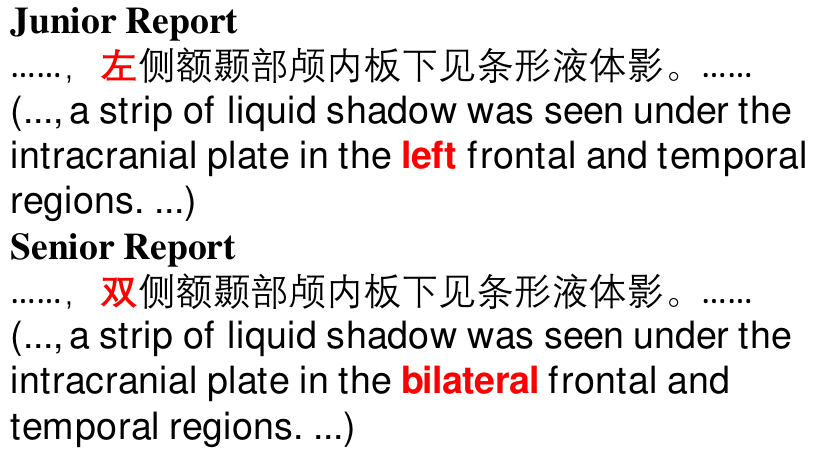}
    \caption{An example pair of junior and senior reports. Although only one Chinese character ``\begin{CJK*}{UTF8}{gbsn}左\end{CJK*}'' (left) is revised by ``\begin{CJK*}{UTF8}{gbsn}双\end{CJK*}''(bi-), they correspond to completely different locations of disease, meaning the junior report is unqualified.}
    \label{example2}
\end{figure}

The basic assumption of STS is to measure the difference degree between any two texts by their global semantics. Unfortunately, for radiology reports, the different degrees can be dominated by locally specific revised tokens, phrases, and n-grams due to the medical specialization, instead of the global semantics. 
As the example shown in Fig.~\ref{example2}, only one Chinese character ``\begin{CJK*}{UTF8}{gbsn}左\end{CJK*}'' (left) is revised by ``\begin{CJK*}{UTF8}{gbsn}双\end{CJK*}''(bi-) in the example pair of junior and senior reports, but they correspond to completely different locations of disease, meaning the junior report is unqualified.


Based on this observation, we instead measure the difference degrees between junior and senior report pairs at the span level. To be specific, we propose \textbf{S}pan-level \textbf{Q}uality \textbf{A}ssurance Evalua\textbf{TOR} (\textbf{\baby}), formulating the QA task as a span-level classification problem. For each junior and senior report pair, we predict the QA scores for the revised spans between them and then integrate the span-level QA scores to form its final QA score. Our \baby consists of two stages, namely report merging processing and span-level classification training. In the first stage, we extract all revised spans between each junior and senior report pair and merge them as a mixed report with the beginning, inside, and outside (BIO)\cite{ramshaw1999text} tags. In the second stage, we apply the pre-trained language models BERT and T5 \cite{devlin2018bert, Raffel2020Exploring} to build a span-level classification model that can predict the QA scores for revised spans. To save many human efforts on span-level labeling, we propose to train the model in a self-training manner.

In this study, we collect 12,013 radiology reports, which cover multiple sections.
The experimental results demonstrate that our proposed \baby can be effective in the QA score prediction task. We also make many qualitative analyses from various perspectives

In summary, we list our contributions as follows:
\begin{itemize}
    \item We formulate the QA task of radiology reports as a span-level classification problem, so as to accurately measure the QA scores at the span level.
    
    \item We propose a novel span-level QA method named \baby, which can be trained in a self-training manner without fine-grained span-level labeling information.
    
    \item We conduct extensive experiments to evaluate the proposed \baby method and make many qualitative analyses from various perspectives.
\end{itemize}


\section{Related Works}
\subsection{Radiology Report Quality Assurance}
A subgroup of studies in this category exploited NLP to assess the quality of content and format of the radiology report itself. As an example of report quality assessment, \cite{gonccalves2022natural} developed a system to automatically identify whether a review of comparison images was properly documented. 
\cite{vosshenrich2021quantifying} investigated whether data mining of quantitative parameters from the report proofreading process can reveal daytime and shift-dependent trends in report similarity as a surrogate marker for resident fatigue. Decreasing report similarity with increasing work hours was observed for day shifts and weekend shifts, which suggests aggravating effects of fatigue on residents’ report writing performances, which improved the training process. \cite{donnelly2019using} used an NLP and machine learning algorithm to evaluate inter-radiologist report variation and compare variation between radiologists using highly structured versus more free-text reporting. The algorithm successfully evaluated metrics showing variability in reporting profiles particularly when there is free text. This variability can be an obstacle to providing effective communication and reliability of care.

\subsection{Pre-trained Language Model}
Pre-trained language models are advanced artificial intelligence systems trained on vast amounts of textual data to understand and generate human-like language. These models employ Transformer\cite{vaswani2017attention} architectures to explore text context, grammar, and meaning. Recently, pre-trained language models have been widely used in traditional NLP fields, such as text classification \cite{yang2019xlnet,li2024graph,mueller2022label, ouyang2022weakly}, sentiment analysis \cite{yang2023s3, tian2020skep,tian2023reducing,wang2022contrastive}, and more. 

With the great success of pre-trained language models, they have also seen a wide range of applications in other domains. In the medical field, pre-trained language models tailored for medical texts are trained specifically on vast repositories of medical literature, clinical notes, disease databases, and other healthcare-related information. 
Pre-training on large-scale data can help physicians better explore the implicit knowledge and connections in medical texts. Nowadays, there are many medical pre-trained language models such as BioBERT~\cite{lee2020biobert}, Med-BERT ~\cite{rasmy2021med}, and more. These models have been widely applied to solve many medical problems such as generating summaries\cite{liu2021medical,liu2021competence,yang2021writing}, extracting information\cite{landolsi2023information, xia2022speaker}, providing a contextual understanding of radiology reports\cite{mrini2022medical,mrini2021gradually}, clinical decision support\cite{yang2022social,zhangLW23a}, EHR management \cite{Zhou2022cancerbert,jade23using,elias23natural}, drug discovery \cite{richard18hybird}, and so on. 




\begin{figure*}[ht]
    \centering
    \includegraphics[width=0.95\textwidth]{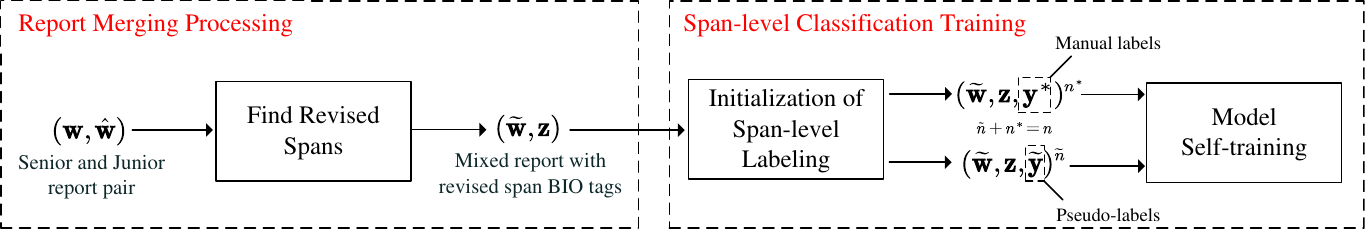}
    \caption{The overall framework of \baby.}  
    \label{framework}
\end{figure*}

\begin{figure*}[t]
    \centering
    \includegraphics[width=\textwidth]{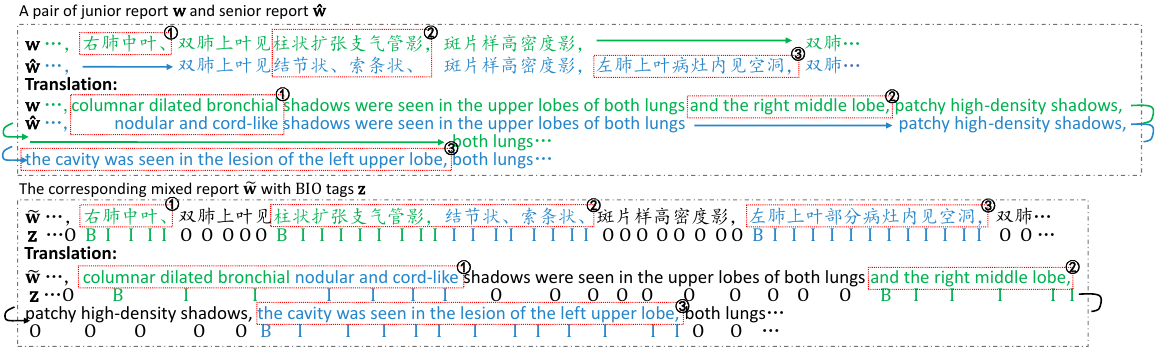}
    
    \caption{An example of the report merging process, which merges a pair of junior and senior reports $(\mathbf{w},\mathbf{\hat w})$ as a mixed report $\mathbf{\widetilde w}$ with BIO tags $\mathbf{z}$. To clearly explain this process, we compare $\mathbf{w}$ and $\mathbf{\hat w}$ by each Chinese character to put all different ones in boxes, and also put all 3 corresponding revised spans of $\mathbf{\widetilde w}$ in boxes. Besides, in $\mathbf{\widetilde w}$, the beginning, inside, and outside of revised spans are marked by $\mathrm{B}$, $\mathrm{I}$, and $\mathrm{O}$, respectively.}    

    \label{examples}
\end{figure*}

\subsection{Span-based NLP Methods}
Span-based neural language processing methods refer to a class of NLP techniques that identify and process spans (i.e. phrases, sentences, or even larger text segments) of text rather than individual tokens (i.e., words or characters). Span-based methods allow for fine-grained analysis by focusing on specific segments of text, enabling detailed linguistic processing. Moreover, these methods streamline processing by narrowing down the focus to relevant spans, reducing computational complexity compared to analyzing entire texts.

Span-based methods are commonly used for named entity recognition \cite{wang2019combining,yu2020improving,li2021a,fu2021spanner,ouyang2024aspect} because the named entities are always text segments rather than single. These models are trained on annotated data to identify and classify spans of text that correspond to named entities. 
Span-based methods are also popular in other sequence labeling tasks~\cite{cheng2023unifying,yang2023spbere}, sentiment analysis~\cite{li2023simple,zhang2023span}, dependency parsing~\cite{gan2022dependency,yang2022headed}, and more.
Following these ideas, we transfer the document-level STS comparison task to the span-label classification task.

\section{Proposed \baby Method}

\subsection{Task and Dataset Description}

\vspace{5pt}
\noindent\textbf{Task definition of QA.} We now briefly describe the task of QA. Formally, we are given a dataset of $n$ labeled samples $\mathcal{D} = \{(\mathbf{w}_i,\mathbf{\hat w}_i, y_i)\}_{i=1}^{n}$, where each of $\mathbf{w}_i$ and $\mathbf{\hat w}_i$ are respectively Chinese radiology reports finished by junior and senior doctors and $y_i \in \{0,1\}$ denotes the corresponding QA score of $\mathbf{w}_i$ evaluated by senior doctors. Specifically, 0/1 implies $\mathbf{w}_i$ is unqualified/qualified. The goal of this task is to induce a classification model over $\mathcal{D}$, which can predict the QA score of any unseen Chinese radiology report pair $(\mathbf{w},\mathbf{\hat w})$.

\vspace{5pt}
\noindent\textbf{Dataset description.}
In this study, we collect a total of 12,013 radiology reports which cover multiple modalities (X-ray, computed tomography, and magnetic resonance imaging) and multiple positions (abdomen, neurology, and chest). 
We categorize the datasets based on positions into four datasets: abdomen, neurology, chest, and full (reports from all three positions) according to the position.
The dataset is divided into 10,811 reports for the training sets and 1,202 reports for the test sets. 
Among these, 10,290 reports are classified as qualified and 1,723 as unqualified.
Given each radiographic image, a junior doctor will write an original report $\mathbf{w}$, formally dubbed as \textbf{junior report}; and then the senior doctor(s) will revise the junior report $\mathbf{w}$ as $\mathbf{\hat w}$, formally dubbed as \textbf{senior report}, and simultaneously evaluate the quality of $\mathbf{w}$ by marking a QA score $y$ for $\mathbf{w}$. In terms of both junior and senior reports, they contain about 150 Chinese characters on average.

\begin{figure*}[t]
    \centering
    \includegraphics[width=0.95\textwidth]{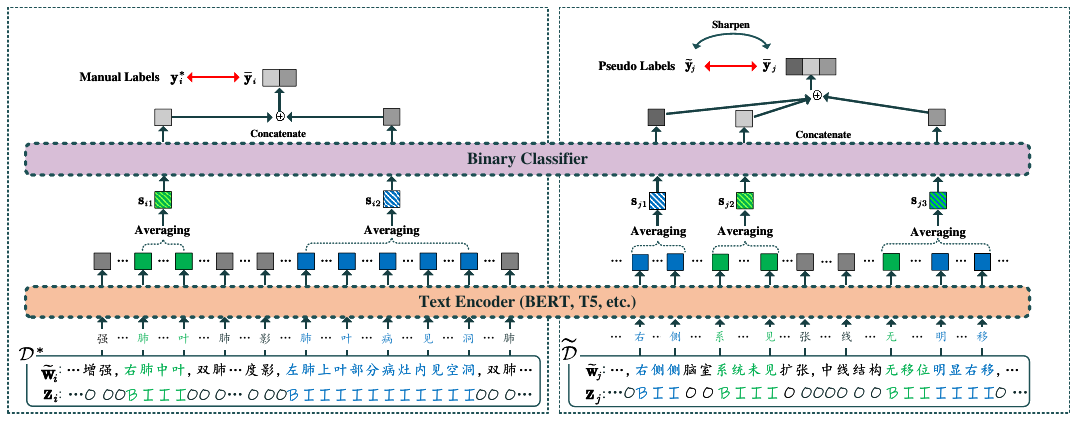}
    \caption{The full process of model self-training.}  
    \label{training}
\end{figure*}

\subsection{Overview of \baby}


Naturally, for each report pair $(\mathbf{w},\mathbf{\hat w})$, a naive way of estimating its QA score is to measure the difference degree between junior and senior versions. Different from the general texts, whose difference degrees are commonly measured by global semantics, the difference degrees between radiology reports can be dominated by locally specific revised tokens, phrases, and n-grams due to the medical specialization. Based on this analysis, we propose to formulate the QA task as a span-level classification problem, where for each pair $(\mathbf{w},\mathbf{\hat w})$, we predict the QA scores for the \textbf{revised spans} between them, and then integrate the span-level QA scores to form its final QA score. To achieve this, we proposed \baby, and as depicted in Fig.\ref{framework}, it consists of two stages, including \textbf{report merging processing} and \textbf{span-level classification training}. We will introduce the two stages in detail in the following sections.

\subsection{Report Merging Processing}

Because the senior report can be treated as a revised version of the junior report, for each pair $(\mathbf{w},\mathbf{\hat w})$, we can merge them as a \textbf{mixed report} $\mathbf{\widetilde w}$ by marking the revised spans between them. To be specific, we automatically extract all different Chinese characters by scanning $(\mathbf{w},\mathbf{\hat w})$ with the longest common sub-sequence algorithm \cite{Hirschbert1977Algorithms} (see Algorithm \ref{alg1}). We then define the revised spans by 3 different types, including \textit{deletion}, \textit{addition}, and \textit{revision}, explained below:

\begin{itemize}
    \item \textbf{Deletion:} The spans are deleted from junior reports by senior doctors.

    \item \textbf{Addition:} The spans are added to senior reports by senior doctors.

    \item \textbf{Revision:} The spans are combined by the ones deleted from junior reports and rewritten in senior reports by senior doctors. 
\end{itemize}

All these 3 kinds of revised spans will be retained in the mixed report $\mathbf{\widetilde w}$, and be marked by BIO\cite{ramshaw1999text} tags $\mathbf{z} \in \{\mathrm{B, I, O}\}^m$, where $\mathrm{B}$, $\mathrm{I}$, and $\mathrm{O}$ are the beginning, inside and outside of revised spans, respectively, and $m$ denotes the total number of Chinese characters in $\mathbf{\widetilde w}$. Accordingly, we can formulate the new span-level sample as $(\mathbf{\widetilde w},\mathbf{z})$ (see example in Fig.\ref{examples}). 


\begin{algorithm}[t]
\caption{Different Chinese character extraction by LCS}
\label{alg1}
\textbf{Input:} A pair of junior and senior reports $(\mathbf{w},\mathbf{\hat w})$

\textbf{Output:} Different Chinese characters $\mathbf{D}$
\begin{algorithmic}
\State Create a matrix $\mathbf M$ with dimensions $\mathrm{len}(\mathbf{w}) \times \mathrm{len}(\mathbf{\hat w})$
\For{$i \gets 0$ to $\mathrm{len}(\mathbf{w})$}
    \For{$j \gets 0$ to $\mathrm{len}(\mathbf{\hat w})$}
        \If{$\mathbf{w}_i == \mathbf{\hat w}_{j} $}
            \State $\mathbf M_{ij} \gets \mathbf M_{(i-1)(j-1)} + 1$
        \Else
            \State $\mathbf M_{ij} \gets \max(\mathbf M_{(i-1)j}, \mathbf M_{i(j-1)})$
        \EndIf
    \EndFor
\EndFor
\State Initialize pointers $i \gets \mathrm{len}(\mathbf{w})$ and $j \gets \mathrm{len}(\mathbf{\hat w})$
\While{$i > 0$ and $j > 0$}
    \If{$\mathbf{w}_i == \mathbf{\hat w}_j$}
        \State Append  $\mathbf{w}_i$ to $\mathbf{L}$
        \State  $i = i - 1$ and $j = j - 1$
    \ElsIf{$\mathbf M_{(i-1)j} \geq \mathbf M_{i(j-1)}$}
        \State $i = i - 1$
    \Else
        \State $j = j - 1$
    \EndIf
\EndWhile
\State $\mathbf{L} \gets$Reverse the list $\mathbf{L}$ to obtain the all common subsequences
\State $\mathbf{D} \gets$ Output the revised Chinese character by removing $\mathbf{L}$ from $\mathbf{w}$ and $\mathbf{\hat w}$

\end{algorithmic}
\end{algorithm}

\subsection{Span-level Classification Training}

After transforming all report pairs $\{(\mathbf{w}_i,\mathbf{\hat w}_i)\}_{i=1}^{n}$ into span-level samples $\{(\mathbf{\widetilde w}_i, \mathbf{z}_i)\}_{i=1}^{n}$, we then train a span-level classification model $\mathcal{F}_{\Theta}(\cdot)$, which can predict the span-level QA scores for revised spans. Unfortunately, we have no such fine-grained span-level labels of QA scores, but only the sample-level labels $\{y_i\}_{i=1}^{n}$. To handle this problem, we treat the problem as a weakly-supervised learning problem and train the model $\mathcal{F}_{\Theta}(\cdot)$ in a self-training manner. Specifically, this stage consists of two steps, including \textbf{initialization of span-level labeling} and \textbf{model self-training}.

\vspace{3pt}
\subsubsection{Initialization of Span-level Labeling}
\label{labeling}
In this step, we expect to initialize the span-level labels of QA scores for revised spans, where, for convenience, we suppose that each span-level sample contains $l$ revised spans. First, we invite 3 senior doctor volunteers to manually assign span-level labels for $n^*$ span-level samples, to form a subset of labeled samples $\mathcal{D}^* = \{(\mathbf{\widetilde w}_i,\mathbf{z}_i,\mathbf{y}^*_i)\}_{i=1}^{n^*}$, where $\mathbf{y}^*_i \in \{0,1\}^l$ and 0/1 denotes the revised span is an unqualified/qualified modification. Because the labor cost of such fine-grained labeling is prohibitive, we only manually generate $n^* = 99$ labeled samples (The number of abdomen, neurology, and chest is 28, 30, and 41 respectively). To save many labor costs, we initialize pseudo-labels for other $\widetilde n = n-n^*$ span-level samples. We are inspired by the assumption that for each report pair if its sample-level label is unqualified/qualified, the corresponding revised spans are likely to be unqualified/qualified. Accordingly, we directly initialize their span-level labels as the corresponding sample-level labels, to form a subset of pseudo-labeled samples $\mathcal{\widetilde D} = \{(\mathbf{\widetilde w}_i,\mathbf{z}_i,\mathbf{\widetilde y}_i)\}_{i=1}^{\widetilde n}$, where $\mathbf{\widetilde y}_i \in \{0, 1\}^l$.

\vspace{3pt}
\subsubsection{Model Self-training}

Given $\mathcal{D}^* \cup \mathcal{\widetilde D}$, we train the span-level classification model $\mathcal{F}_{\Theta}(\cdot)$ in a self-training manner. As depicted in Fig.~\ref{training}, the model $\mathcal{F}_{\Theta}(\cdot)$ is composed of a text encoder $\mathcal{E}_{\Phi}(\cdot)$ and a binary classifier $\mathcal{C}_{\mathbf{W}}(\cdot)$, \ie $\Theta = \{\Phi,\mathbf{W}\}$. Here, we apply the pre-trained language models BERT \cite{devlin2018bert} and T5 \cite{Raffel2020Exploring} as the text encoders, which ingest each mixed report $\mathbf{\widetilde w}_i$ and generate the embeddings $\mathbf{H}_i$ of all its Chinese characters:
\begin{equation}
    \mathbf{H}_i = \mathcal{E}_{\Phi}(\mathbf{\widetilde w}_i).
\end{equation}

Accordingly, each of the $j$-th revised spans of $\mathbf{\widetilde w}_i$ can be represented by the average embedding of the corresponding Chinese characters marked by $\mathbf{z}_i$:
\begin{equation}
    \mathbf{s}_{ij} = \frac{1}{|\mathbf{g}_j|} \sum _{g \in \mathbf{g}_j } \mathbf{H}_{ig},
\end{equation}
where $\mathbf{g}_j$ denotes the character indexes of the $j$-th span, starting with the character index of $j$-th $\mathrm{B}$ tag and ending with the character index before the next $\mathrm{O}$ tag or $\mathrm{B}$ tag.

Given those embeddings of revised spans, the binary classifier is used to predict their important scores as follows:
\begin{equation}
    \bar{y}_{ij} = \mathcal{C}_{\mathbf{W}}(\mathbf{s}_{ij}).
\end{equation}

Then, we update the model parameters $\Theta$ with the labels obtained in Section.~(\ref{labeling}). For the reports in $\mathcal{D}^*$, whose labels are given by the senior doctor. We use them to directly train the model by minimizing the following equation:
\begin{equation}
    \mathcal{L}_{manual} = \frac{1}{n^*}\sum_{i\in \mathcal{D}^*}\ell(\bar{\mathbf{y}}_i, \mathbf{y}_i^*),
\end{equation}
where $\ell$ denotes the cross-entropy loss and $\bar{\mathbf{y}}_i = \{\bar{y}_{ij}\}_{j=1}^l$. 

For the reports in $\widetilde{\mathcal{D}}$ which are pseudo-labeled, we first update $\Theta$ and then refine the pseudo-labels by the model predictions. Similarly, we update $\Theta$ by minimizing:
\begin{equation}
    \mathcal{L}_{pseudo} = \frac{1}{\widetilde{n}}\sum_{i\in \mathcal{\widetilde{D}}}\ell(\bar{\mathbf{y}}_i, \widetilde{\mathbf{y}}_i).
\end{equation}
With the updated span classifier, we re-predict the important scores $\hat{\mathbf{y}}_i=\{\bar{y}_{ij}\}_{j=1}^l$ for revised spans in $\mathbf{\widetilde w}_i$. Then we update the pseudo-labels with $\hat{\mathbf{y}}_i$, forming new pseudo-labels for the revised spans. The pseudo-labels are updated by the following equation:
\begin{equation}
    \tilde{y}_{ij}=
\begin{cases}
	\tilde{y}_{ij} \;\;\;\; l_{ij}\geq\gamma ,\\
	\bar{y}_{ij} \;\;\;\; l_{ij}<\gamma ,\\
\end{cases}
\label{eq_gamma}
\end{equation}
where $l_{ij}$ denotes the loss value with respect to $\{\mathbf{s}_{ij}, \tilde{y}_{ij}\}$.

Overall, we update the model parameter $\Theta$ by minimizing:
\begin{equation}
    \mathcal{L}_{all} =\mathcal{L}_{manual} + \lambda \mathcal{L}_{pseudo},
\label{eq_lambda}
\end{equation}
here, $\lambda$ is a hyper-parameter that adjusts the weight of pseudo-labels in the training process to account for noises in the pseudo-labels.

\begin{table*}
\caption{Experimental results of our \baby and other baselines on four datasets}
\centering
\small
\renewcommand\arraystretch{1.1}
\begin{tabular}{p{2.5cm}*{16}{p{0.55cm}}}
\hline
\hline
                &\multicolumn{4}{c}{Abdomen}                                        &\multicolumn{4}{c}{Neurology}                                      &\multicolumn{4}{c}{Chest}                                          &\multicolumn{4}{c}{Full}    \\\midrule 
{Method}        &Acc.            &F1              &Pre.            &Rec.            &Acc.            &F1              &Pre.            &Rec.            & Acc.           & F1             &Pre.            &Rec.            & Acc.           & F1             &Pre.            &Rec.            \\ \midrule   
{Llama3-8B}     & 90.05          & 79.23          & 77.84          & 80.89          & 91.75          & 75.72          & 81.76          & 72.07          & 91.00          & 80.11          & 79.17          & 81.14          & 90.65          & 74.56          & 79.21          & 71.53 \\\midrule 
T5-Cosine       & 88.06          & 53.93          & 53.78          & 84.08          & 88.75          & 47.02          & 44.37          & 50.00          & 87.00          & 48.37          & 56.31          & 50.57          & 87.98          & 48.15          & 60.79          & 50.51          \\ 
{T5-Concat}     & \textbf{92.54} & 79.63          & 74.78          & \textbf{88.57} & 87.50          & 61.83          & 65.81          & 59.97          & 85.50          & 70.76          & 68.83          & 73.71          & 82.64          & 65.53          & 63.72          & 69.12          \\
T5-Abs         & 91.54          & 79.13          & 76.72           & 82.32          & 89.75          & 68.31          & 74.84          & 65.12          & 89.50          & 62.16          & \textbf{89.74} & 58.86          & 88.48          & 69.82          & 72.05          & 68.17          \\ 
\baby-T5-Ave.  & 92.25         & \textbf{80.50}  & \textbf{84.74} & 77.44          & \textbf{90.75} & \textbf{77.06} & \textbf{76.81} & 77.32          & \textbf{91.75} & \textbf{82.20} & 80.56          & 84.14         & \textbf{88.83} & 71.33          & \textbf{73.50} & 69.67          \\ 
\baby-T5-Min.   & 91.25          & 79.91          & 80.19          & 79.64          & 89.25          & 75.22          & 73.47          & \textbf{77.45} & 90.00          & 80.39          & 77.06          & \textbf{85.71}& 88.42          & \textbf{73.21} & 72.74          & \textbf{73.70}  \\\midrule 
Bert-Cosine     & 87.56          & 48.60          & \textbf{93.77} & 50.98          & 88.75          & 47.02          & 44.37          & 50.00          & 87.50          & 46.67          & 43.75          & 50.00          & 88.15          & 46.85          & 44.07          & 50.00
\\
{Bert-Concat}   & 87.81          & 74.39          & 76.26          & 72.91          & 88.00          & 65.91          & 68.64          & 64.13          & 84.00          & 69.86          & 67.46          & 74.57          & 88.73          & 73.12          & 73.05          & 73.19          \\
Bert-Abs        & 90.30          & 78.29          & 78.52          & 78.05          & 88.25          & 67.72          & 69.76          & 66.21          & 86.25          & 72.07          & 70.09          & 75.00          & 84.72          & 63.34          & 63.38          & 63.30          \\
\baby-Bert-Ave. & 94.25          & 87.22          & 87.57          & \textbf{86.87} & \textbf{95.00} & \textbf{87.72} & 87.02          & \textbf{88.45} & \textbf{92.00} & 82.32          & \textbf{81.31} & 83.43          & 91.83          & 77.94          & 83.36          & 74.43          \\  
\baby-Bert-Min. & \textbf{94.75} & \textbf{87.49} & 91.48          & 84.39          & \textbf{95.00} & 86.70          & \textbf{89.23} & 84.57          & 91.75          & \textbf{83.00} & 80.26          & \textbf{86.71} & \textbf{92.33} & \textbf{79.90} & \textbf{84.19} & \textbf{76.84} \\ 
                                                   
\hline 
\hline
\end{tabular}
\label{results}
\end{table*}

\begin{figure*}[t]
    \centering
    \includegraphics[width=\textwidth]{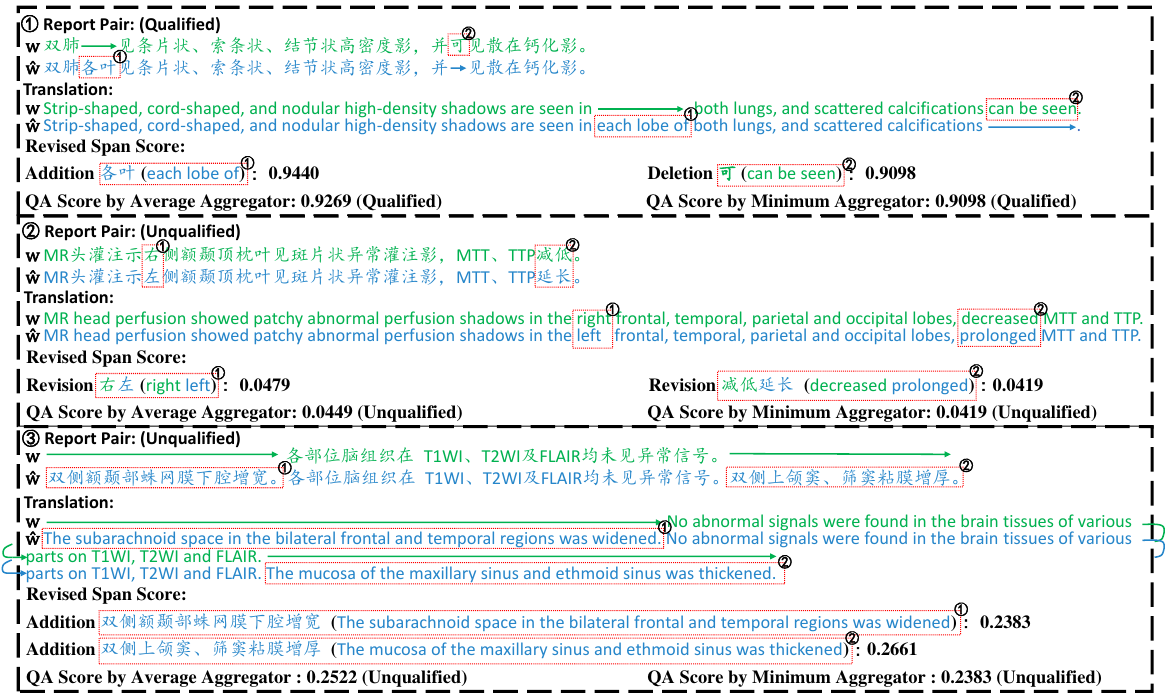}
    \caption{We illustrate several examples for the case study. The revised report is the junior report modified according to the senior report. The revised span score is the important score of each revised span predicted by the model. The QA score is calculated by two aggregators through revised span scores.}
    \label{case study}
\end{figure*}

\subsection{Revision Important Scores Aggregation}

Given the revision important scores $\bar{\mathbf{y}}_i = \{\bar{y}_{ij}\}_{j=1}^l$, predicted through classifier, we propose two aggregation methods to combine these scores for classifying the quality of the junior reports: the average aggregator (\textbf{Ave.}) and the minimum aggregator (\textbf{Min.}).
For the average aggregator, the QA score $y_i$ is calculated as:
\begin{equation}
y_i = \frac{1}{l}\sum_{j=1}^{l}\bar{y}_{ij},
\end{equation}
indicating that the QA score is determined by the average importance score across all revised spans.
For the minimum aggregator, the QA score $y_i$ is calculated as:
\begin{equation}
y_i = \min_{1\leq j\leq l}\bar{y}_{ij},
\end{equation}
implying that the report is considered unqualified if at least one revised span is predicted to be unqualified.

\section{Experiments}

\begin{figure*}[t]
    \centering
    \includegraphics[width=0.95\textwidth]{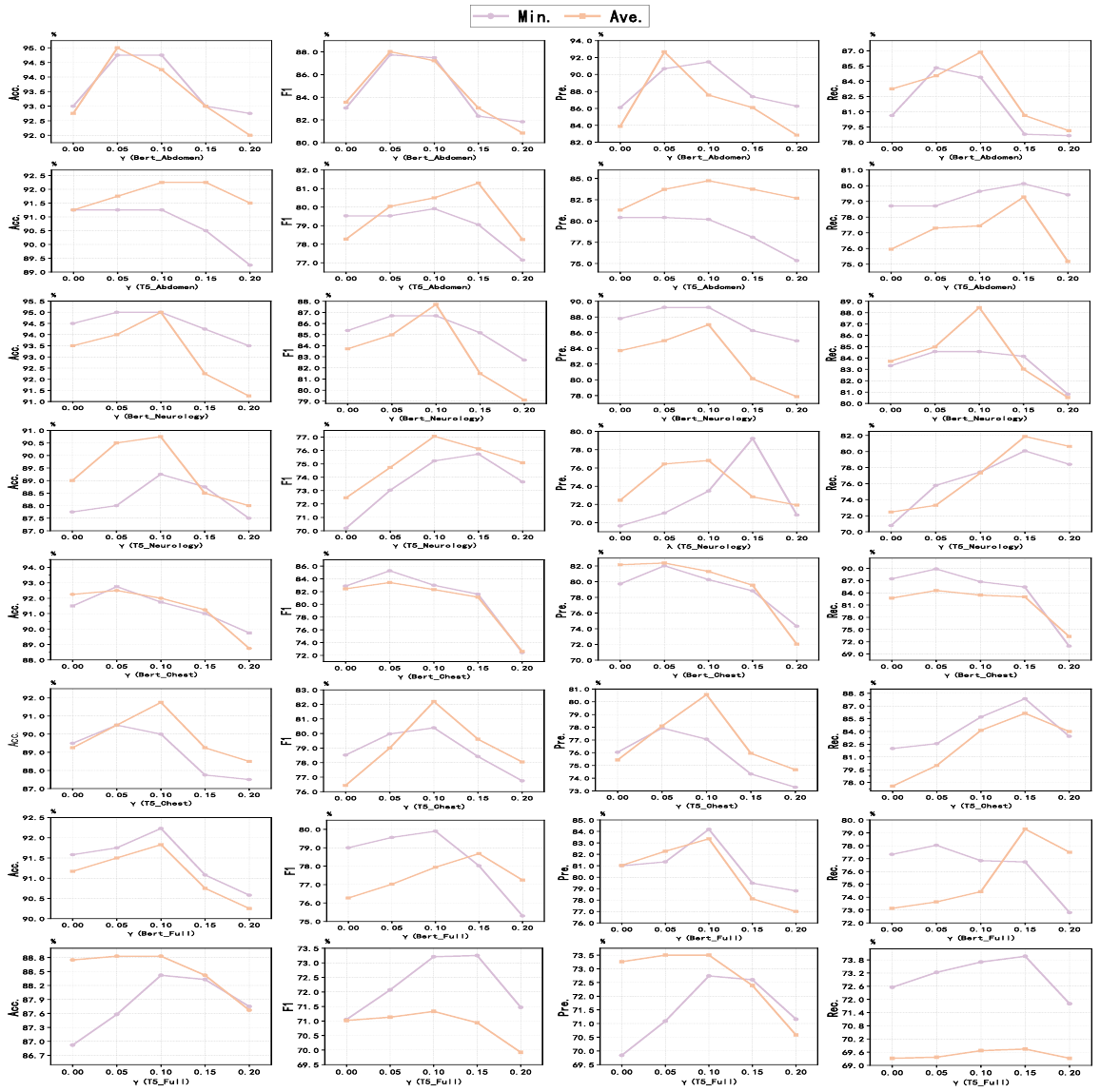}
    \caption{The sensitivity analysis of the $\gamma$ parameter.}  
    \label{gamma sensitivity}
\end{figure*}

\begin{figure*}[t]
    \centering
    \includegraphics[width=0.95\textwidth]{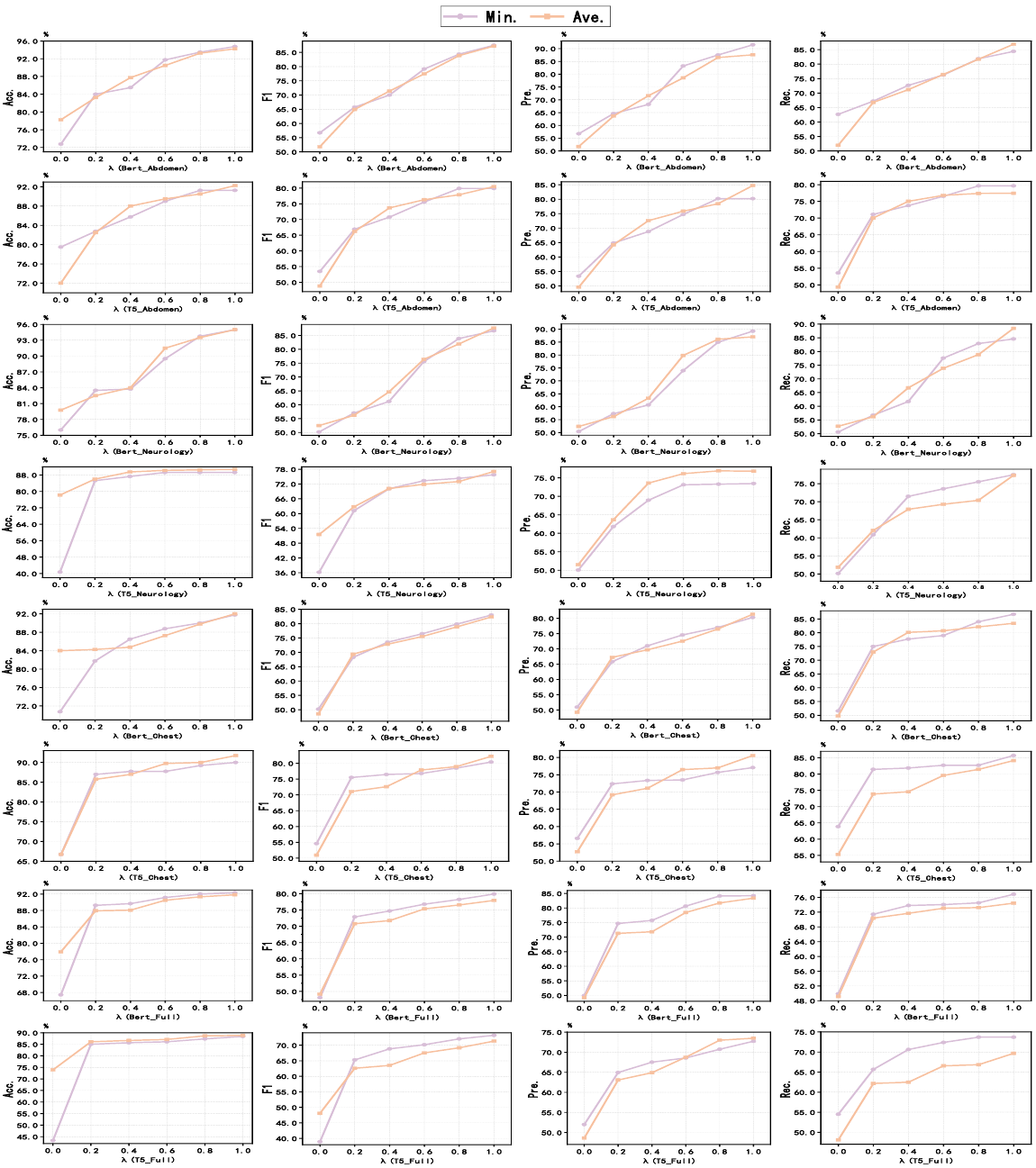}
    \caption{The sensitivity analysis of the $\lambda$ parameter.}  
    \label{lambda sensitivity}
\end{figure*}

\subsection{Baseline and Implementation Details}
\subsubsection{Baseline Models}
In the experiment, we employ three kinds of baseline models listed below:
\begin{itemize}
    \item \textbf{LLM-IFT baseline}: We employ the commonly used large language model Llama3-8B,\footnote{Meta-Llama3-8B-Instruct available at https://huggingface.co/meta-llama/Meta-Llama-3-8B-Instruct} and further update it to predict QA scores by using the labeled reports with instruction fine-tuning.

    \item \textbf{Text similarity baselines}: We regard the QA task as a text similarity between junior and senior reports. We employ two text encoders Bert\footnote{Chinese-bert-wwm-ext available at https://huggingface.co/hfl/chinese-bert-wwm-ext} and T5\footnote{mengzi-t5-base available at https://huggingface.co/Langboat/mengzi-t5-base} and use the cosine similarity as the metric. We fine-tune Bert and T5 over the labeled reports to achieve two baselines, namely \textbf{Bert-Cosine} and \textbf{T5-Cosine}, respectively.

    \item \textbf{Classifier baselines}: We formulate the QA task as a classification problem. We also employ Bert and T5 as the text encoders and use an MLP layer as the classifier. We form the fused embeddings for classification using two techniques. The first is to concatenate the embeddings of junior and senior reports, leading to two baselines \textbf{Bert-Concat} and \textbf{T5-Concat}, respectively. The other is to form the difference between the embeddings of junior and senior reports by absolute values, leading to two baselines \textbf{Bert-Abs} and \textbf{T5-Abs}, respectively.   
\end{itemize}

\subsubsection{Implementation Details}
For \baby, Text similarity methods, and Classifier methods, we use the Adam optimizer with a learning rate of $1\times10^{-6}$ for the text encoder.
In \baby and the Classifier methods, the classifier's learning rate is set to $1\times10^{-3}$.
We employ the OTSU algorithm\cite{otsu1975threshold} to control the classifier's threshold.
The gradient accumulation step is set to 2 for the method using the T5 encoder on neurology datasets and other comparative methods utilize a gradient accumulation step of 1.
For the LLM-IFT method, we use the AdamW optimizer with a learning rate of $1\times10^{-4}$ and set the gradient accumulation step to 8.
We utilize low-rank adaption (LORA)\cite{DBLP:conf/iclr/HuSWALWWC22} to train the model and the rank of LORA is set to 8.
The prompt for the LLM-IFT method is ``\begin{CJK*}{UTF8}{gbsn}根据第一个放射线报告，判断第二个放射线报告是否合格，合格输出‘是’，不合格输出‘否’\end{CJK*}'' (``According to the first radiology report, determine if the second radiology report is qualified. If qualified, output `yes', if not qualified, output `no'"). The Llama3-8B model is constrained to output only one word.
All methods are trained for 100 epochs with a batch size of 8.

To evaluate the performance, we employ several commonly used metrics including Macro-Accuracy (Acc.), Macro-Precision (Pre.), Macro-Recall (Rec.), and Macro-F1 (F1) \cite{Grandini2020Metrics}.

\subsection{Main Results}

TABLE \ref{results} presents the comparative results of three types of baselines and our two proposed methods on four datasets.
We highlight the highest performance for each pre-trained language model across all datasets in bold.
In most evaluation metrics, the \baby method consistently outperforms the other methods.
Specifically, for methods using T5 as the text encoder, \baby achieves improvements in F1 scores of 0.87, 8.75, 11.44, and 3.39 across four datasets compared to the highest-performing baseline.
Similarly, for methods using Bert as the text encoder, \baby achieves gains of 9.20, 20.00, 10.93, and 6.78 in F1 scores over the baselines.
When comparing the \baby method with the LLM-IFT baseline, even though the large language model has more parameters and greater fitting capacity, the proposed span-level evaluator surpasses this document-level semantic classification method.
Due to the imbalance between qualified and unqualified reports, methods such as Bert-Cosine and T5-Cosine achieve high Pre. but low F1 scores.
This indicates that such models tend to classify most reports as qualified, failing to achieve high F1 scores.
In contrast, the \baby method demonstrates better handling of class imbalance, achieving superior F1 scores while balancing data across different categories.

Notably, methods using Bert as the text encoder outperform those using T5, suggesting that Bert, which is designed for encoding tasks, is better suited for classifying information extracted from text through the \baby than T5, which is optimized for generative tasks.
Lastly, the performance differences between the average and minimum aggregators vary across datasets, suggesting that different data types are suitable for different aggregator strategies.


\subsection{Case Study}

We conduct a case study, as illustrated in Fig.\ref{case study}.
We present three representative cases: a qualified case and two unqualified cases.
For each pair of reports, we list the revised spans, the revised span score, and the QA score.
In the first qualified case, the model assigns a score of 0.9440 to the addition of ``\begin{CJK*}{UTF8}{gbsn}各叶\end{CJK*}'' (each lobe of) and 0.9098 to the deletion of ``\begin{CJK*}{UTF8}{gbsn}可\end{CJK*}'' (can be seen).
Both revisions affect only the stylistic expression of the report without changing its content.
The average and minimum aggregators both correctly classify this report as qualified.
In the last two unqualified cases, the model assigns scores of 0.0479, 0.0419, 0.2383, and 0.2661 to the four spans, respectively. 
All the revisions modify the meaning of reports.
Both aggregators correctly evaluate that these reports are unqualified.

We invite a clinician to evaluate the cases and observe the following.
The first case contains the addition of ``\begin{CJK*}{UTF8}{gbsn}各叶\end{CJK*}'' (each lobe of) and the deletion of ``\begin{CJK*}{UTF8}{gbsn}可\end{CJK*}'' (can be seen), which are deemed non-essential for diagnosis or decision-making. 
The second case includes modifications from ``\begin{CJK*}{UTF8}{gbsn}右\end{CJK*}'' (right) to ``\begin{CJK*}{UTF8}{gbsn}左\end{CJK*}'' (left), convey an opposite meaning.
Another revision changes ``\begin{CJK*}{UTF8}{gbsn}减低\end{CJK*}'' (decreased) to ``\begin{CJK*}{UTF8}{gbsn}延长\end{CJK*}'' (prolonged), reflecting differences in measurement angles, namely signal size and time length.
These mistakes, attributed to a doctor's oversight, could result in severe consequences.
The third case involves a missed diagnosis, which could affect clinical decision-making and cause incomplete treatment options.
The clinician thinks that the revisions in the third case are less significant than those in the second case, aligning with the model's evaluation.

\subsection{Sensitivity Analysis}
In our method, there are two crucial parameters, $\gamma$ of $\tilde{y}_{ij}$ and $\lambda$ of $\mathcal{L}_{all}$, which are defined in Eqs. \ref{eq_gamma} and \ref{eq_lambda}, respectively.
This section explores the sensitivity of these parameters to demonstrate their impact on the performance of \baby.
We conduct experiments on all four datasets.
The experimental results are visualized in Fig. \ref{gamma sensitivity} and Fig. \ref{lambda sensitivity}.
The evaluation metrics include Acc., Pre., Rec., and F1.
Experiments are performed with $\gamma$ values of \{0.00, 0.05, 0.10, 0.15, 0.20\} and $\lambda$ values of \{0.0, 0.2, 0.4, 0.6, 0.8, 1.0\}, with only the analyzed parameters varying while all others remain constant.

The results indicate that the best performance is achieved when $\gamma$ lies in the range of \{0.05, 0.10, 0.15\}. A small $\gamma$ leads to a low pseudo-label update frequency, which disrupts classifier training.
Conversely, a large $\gamma$ results in an excessively high update frequency, preventing the classifier from effectively learning the pseudo-label information.
For $\lambda$, the model achieves its highest performance when $\lambda$ equals 1.0.
As $\lambda$ decreases, the model's performance progressively declines, with the worst results observed when $\lambda$ is 0.0.
This indicates that the model effectively leverages information from pseudo-labels.

\section{Conclusion}
In this paper, we address the QA issue in radiology reports by proposing a span-level QA evaluator named \baby to evaluate whether the junior reports are qualified automatically.
Specifically, \baby merges junior and senior reports by marking the revised spans and predicts the qualification of junior reports based on the importance scores of these spans.
To achieve this, we introduce a self-training manner that eliminates the need for fine-grained, span-level labeled reports.
Experimental results demonstrate that \baby significantly outperforms baseline methods.
Additionally, the important score predicted by \baby is consistent with the judgment of senior doctors.

\section*{Acknowledgement}
We would like to acknowledge support for the National Natural Science Foundation of China (No.62276113), the Changchun Science and Technology Bureau Project 23YQ05.

\section*{GenAI Usage Disclosure}
Generative AI tools (ChatGPT) were used only to edit and improve the quality of our existing text to correct spelling and grammar. All ideas, analyses, and results are the authors’ own.

\balance
\bibliography{ref}

\end{document}